\documentclass[preprint, 5p, 10pt]{elsarticle}

\usepackage{amssymb}

\usepackage{amsmath,amsfonts}
\usepackage{algorithmic}
\usepackage{algorithm}
\usepackage{multirow}
\usepackage{graphicx}
\usepackage{hyperref}
\hypersetup{
  colorlinks,
  urlcolor=blue,
  linkcolor=blue,       
  anchorcolor=blue,  
  citecolor=blue,        
}

\journal{Information Fusion}

\begin{document}
\sloppy

\begin{frontmatter}



\title{Adversarial Purification of Information Masking}


\author[njust]{Sitong Liu}
\ead{lstnjust@163.com}
\author[njust]{Zhichao Lian\corref{cor1}}
\ead{lzcts@163.com}
\author[njust]{Shuangquan Zhang}
\ead{zhangsq@njust.edu.cn}
\author[njust]{Liang Xiao\corref{cor1}}
\ead{xiaoliang@njust.edu.cn}

\cortext[cor1]{Corresponding author}
\address[njust]{organization={School of Computer Science and Engineering, Nanjing University of Science and Technology},
            addressline={No.200 Xiaolingwei Street}, 
            city={Nanjing},
            postcode={210094}, 
            state={Jiangsu},
            country={China}}

\begin{abstract}
Adversarial attacks meticulously generate minuscule, imperceptible perturbations to images to deceive neural networks. Counteracting these, adversarial purification methods seek to transform adversarial input samples into clean output images to defend against adversarial attacks. Nonetheless, extent generative models fail to effectively eliminate adversarial perturbations, yielding less-than-ideal purification results. We emphasize the potential threat of residual adversarial perturbations to target models, quantitatively establishing a relationship between perturbation scale and attack capability. Notably, the residual perturbations on the purified image primarily stem from the same-position patch and similar patches of the adversarial sample. We propose a novel adversarial purification approach named Information Mask Purification (IMPure), aims to extensively eliminate adversarial perturbations. To obtain an adversarial sample, we first mask part of the patches information, then reconstruct the patches to resist adversarial perturbations from the patches. We reconstruct all patches in parallel to obtain a cohesive image. Then, in order to protect the purified samples against potential similar regional perturbations, we simulate this risk by randomly mixing the purified samples with the input samples before inputting them into the feature extraction network. Finally, we establish a combined constraint of pixel loss and perceptual loss to augment the model's reconstruction adaptability. Extensive experiments on the ImageNet dataset with three classifier models demonstrate that our approach achieves state-of-the-art results against nine adversarial attack methods. Implementation code and pre-trained weights can be accessed at \textcolor{blue}{https://github.com/NoWindButRain/IMPure}.
\end{abstract}

\onecolumn

\begin{graphicalabstract}
\includegraphics[width=\linewidth]{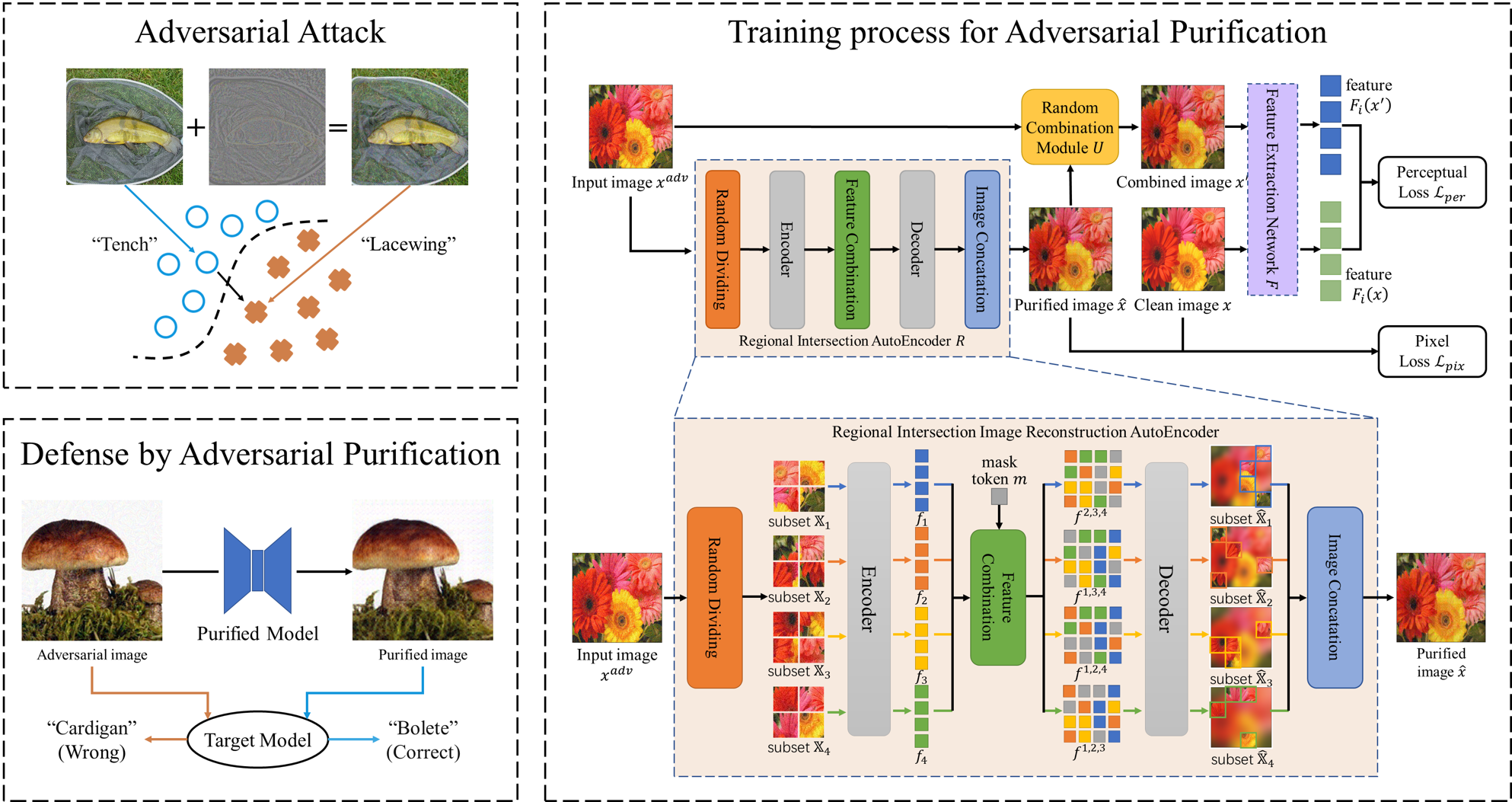}
\end{graphicalabstract}

\begin{highlights}
\item Establishes a quantitative correlation between residual adversarial perturbation scale and attack capability.
\item Introduces various information masking strategies to resist the same-position adversarial perturbation.
\item Advances a regional intersection reconstruction approach aimed at diminishing the scale of residual same-position adversarial perturbations.
\item Proposes a simulated technique of residual perturbation to fortify defenses against content-similar adversarial perturbations.
\item Explores a joint constraint form encompassing pixel loss and perceptual loss to augment the flexibility in clean sample generation.
\end{highlights}

\twocolumn


\end{frontmatter}


\section{Introduction}
\label{sec:intro}

\begin{figure}[htb]
  \centering
  \includegraphics[width=0.9\linewidth]{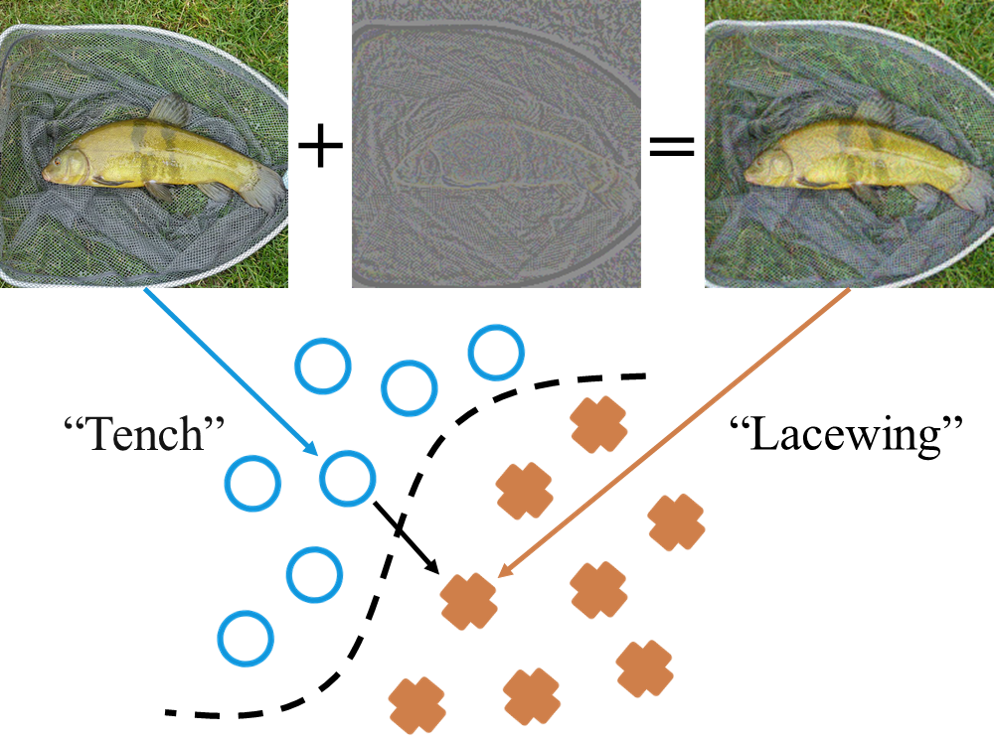}
  \caption{Adversarial Attack Process: The illustration depicts a pair of images—the pristine image on the left is accurately classified as a tench by the target model. Conversely, the image on the right showcases an adversarial example, derived post the infusion of a minor perturbation. This adversarial perturbation alters the feature mapping, propelling it beyond the decision boundary delineated by the target model, and consequently, the adversarial sample is misclassified as a lacewing.}
  \label{fig:AdvProcess}
\end{figure}

Neural network-based machine learning methodologies have become instrumental in a multitude of domains, including autonomous driving~\cite{Prakash2021MultiModalFT} and access control systems~\cite{Huang2021WhenAF}. Nonetheless, emerging research has revealed the vulnerability of neural networks to adversarial attacks~\cite{Goodfellow2014ExplainingAH}. As can be seen from Fig. \ref{fig:AdvProcess}, even slight perturbations can induce erroneous outputs from the target model, diverging from original sample predictions. Adversarial attack strategies possess not only the capability to deceive neural networks but also transferable applicability, capable of inflicting damage on black-box models without the requisite knowledge of the target model’s architecture or parameters~\cite{Papernot2016PracticalBA, Chen2017ZOOZO, Brendel2017DecisionBasedAA}. Given the significant security vulnerabilities these adversarial attacks introduce, the development of efficient defense mechanisms becomes vital, aiming to enhance the adversarial robustness of neural networks.

One of the primary objectives of defensive methods is to ensure the target model’s immunity to adversarial perturbations, maintaining accurate outputs, especially in high-stakes applications like autonomous driving and access control. Defenses can generally be categorized into model-specific and model-agnostic strategies~\cite{Guo2018CounteringAI}. Model-specific methods, including adversarial training~\cite{Goodfellow2014ExplainingAH} and gradient masking~\cite{Dhillon2018StochasticAP}, augment the robustness against adversarial attacks by modifying the training strategy of the target model. However, Model-specific methods have notable downsides like a negative robustness-accuracy correlation~\cite{Tsipras2018RobustnessMB} and the necessity of retraining, which can be resource-intensive. Model-agnostic methods, such as JPEG compression~\cite{Dziugaite2016ASO}, random padding~\cite{Xie2017MitigatingAE}, and high-level representation guided denoising (HGD)~\cite{Liao2017DefenseAA}, operate at the input stage, preprocessing input samples to diminish adversarial perturbations' impact. Model-agnostic methods, while intuitive and seamlessly integrated, often struggle to entirely clean adversarial perturbations and preserve image quality, while also being potentially susceptible to compromise in a white-box setting.

\begin{figure}[htb]
  \centering
  \includegraphics[width=\linewidth]{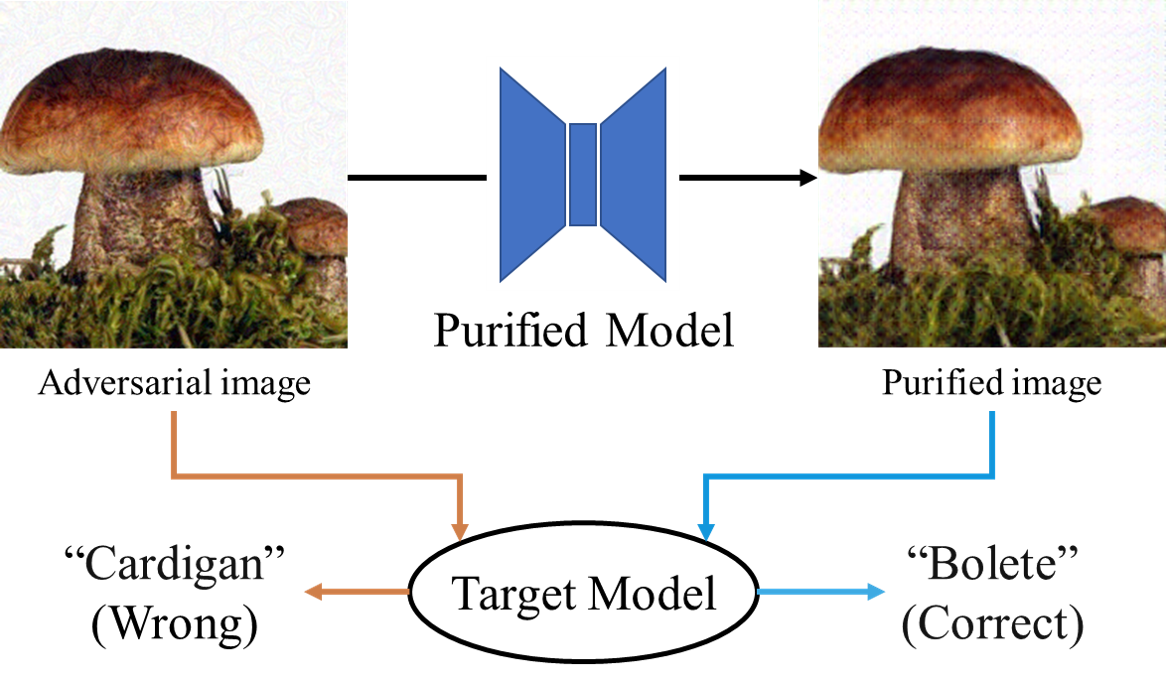}
  \caption{Adversarial Attack Process: The illustration depicts a pair of images—the pristine image on the left is accurately classified as a tench by the target model. Conversely, the image on the right showcases an adversarial example, derived post the infusion of a minor perturbation. This adversarial perturbation alters the feature mapping, propelling it beyond the decision boundary delineated by the target model, and consequently, the adversarial sample is misclassified as a lacewing.}
  \label{fig:AdvPuriProcess}
\end{figure}

Generative model-based adversarial purification methods are an important part of model-agnostic methods. Generative models have seen extensive deployment across various computer vision and deep learning tasks, including image denoising~\cite{Zhang2022PracticalBI} and image restoring~\cite{Chen2022SimpleBF}. As shown in Fig. \ref{fig:AdvPuriProcess}, Viewing adversarial perturbations as a distinct type of noise, researchers have proposed adversarial purification methods using generative models to denoise or reconstruct adversarial samples prior to input into the target model. However, the pixel-level loss functions, typically employed in traditional generative tasks, fall short in adversarial purification due to the error amplification effect. Some approaches~\cite{Liao2017DefenseAA, Zhang2021DefenseAA} employ perceptual loss functions, calculating feature differences in target model intermediate layers as a remedy. It's noteworthy that the noise scale in purified images might exceed that of the original adversarial images, not only degrading image quality but also potentially undermining the accuracy of the target model. In summary, our adversarial purification methodology is structured around three primary objectives: 1) To extensively eliminate adversarial perturbations present in the input image; 2) To effectively defend against any residual adversarial perturbations; 3) To minimize the discrepancy between the purified image and its original, clean counterpart.

In this paper, we delve into the detrimental impact of residual adversarial perturbations on purified images, furnishing rigorous proof that articulates the quantitative relationship between the scale of residual perturbations and the capabilities of the attack. Consequently, we emphasize the importance of limiting the scale of adversarial perturbations for adversarial purification methods. The residual perturbations observed on the purified image patches come from same-position patches and content-similar patches in the adversarial image. We propose a novel adversarial purification method, termed Information Mask Purification (IMPure), devised to resist same-position and content-similar perturbations through the meticulous design of a Regional Intersection image reconstruction Autoencoder (RIAE) and a Random Combination Module (RCM), respectively. RIAE is based on the Masked Autoencoder (MAE)~\cite{He2021MaskedAA} and showcases robust reconstruction capabilities through a reconfigured and parallelized architecture. We selectively mask part of the adversarial sample patch information, using the remaining patches to reconstruct the masked patch and parallelizing the reconstruction process on all patches to obtain a complete purified image. Given the minute pixel values of the adversarial perturbation, the influence of the perturbation can be effectively mitigated through information masking. The image information loss of the masked patches can be reconstructed by leveraging the information from the remaining intact patches through the reconstruction network. Identifying content-similar patches poses a challenge, yet RCM achieves resistance to content-similar perturbations indirectly. Operational during the training phase, RCM strategically merges the purified image with the adversarial image, deliberately preserving some adversarial perturbations. Thus, to counteract the impact of preservational perturbations on the target model, the reconstruction network needs to amplify the overall defense capability of the purified image. Furthermore, our proposed method proffers enhanced flexibility in selecting feature extraction networks relative to existing adversarial purification methods. We employ joint constraints of pixel loss and perceptual loss, facilitating the integration of both low-level pixel information and high-level perceptual information throughout the training phase, thereby helping the model on a virtuous cycle.

Our work has made the following main contributions:

\begin{enumerate}[1.]
  \item We meticulously explore the perils posed by residual adversarial perturbations within adversarial purification strategies, elucidating a quantitative nexus between perturbation scale and offensive capabilities.
  \item We propose IMPure, an innovative adversarial purification method, which consists of RIAE that maximizes the elimination of same-position perturbations and RCM that encourages the network to improve the defense capabilities to resist content-similar perturbations. Additionally, we provide a flexible joint-constrained loss to promote the consistency of purified images in both low-level information and high-level representation.
  \item A suite of comprehensive experiments demonstrates that our method achieves excellent defense performance on three target networks and nine adversarial attack methods in the classification task of the ImageNet dataset.
\end{enumerate}

\section{Related works}

In this section, we introduce representative methods for adversarial attack and defense. To simplify the problem, we focus only on the image classification task. First, we designate some notations used in this paper. Let $\boldsymbol{x}$ denote the original clean image from a given dataset, and $y$ denotes the corresponding label. A neural network $f:\boldsymbol{x}\rightarrow y$ is called the target model. $\mathcal{L}(\boldsymbol{x}, y)$ denotes the loss function of the network. $p(y|\boldsymbol{x})$ predicted probability of class $y$ output by the network $f$. $y_{\boldsymbol{x}}=\mathop{\arg\max}_{y}p(y|\boldsymbol{x})$ is the predicted class of $\boldsymbol{x}$. $\boldsymbol{x}^{adv}$ denotes an adversarial example of $\boldsymbol{x}$.

\subsection{Adversarial attack}

The adversarial attack method generates an adversarial sample $\boldsymbol{x}^{adv}$ on a clean image by carefully constructing an adversarial perturbation that causes the classifier $f$ to give wrong predictions $y_{\boldsymbol{x}^{adv}}$. Based on their purpose and environment setting respectively, adversarial attack methods are categorised, and some representative methods are introduced in this section.

Adversarial attacks can be divided into targeted and untargeted attacks according to their purposes. The goal of an untargeted attack is $y_{\boldsymbol{x}^{adv}} \neq y$. The target attack specifies a specific class $y^{\prime}$ and makes $y_{\boldsymbol{x}^{adv}} = y^{\prime}$. 

Goodfellow et al.~\cite{Goodfellow2014ExplainingAH} suggested the cumulative effects of high dimensional model weights and the deep model is more linear in high dimensions. They proposed an untargeted method called the Fast Gradient Sign Method (FGSM) to generate adversarial examples only by computing the gradient once:
\begin{equation}
  \boldsymbol{x}^{adv}=\boldsymbol{x}+\epsilon sign(\bigtriangledown_{\boldsymbol{x}} \mathcal{L}(\boldsymbol{x},y))
  \label{eq:fgsmloss}
\end{equation}%
where $\epsilon$ is the strength of the perturbation, $sign(\cdot )$ represents a symbolic function. By modifying the FGSM to maximize the probability of a specific class $y^{target}$, a version of the target attack can be obtained:
\begin{equation}
  \boldsymbol{x}^{adv}=\boldsymbol{x}-\epsilon sign(\bigtriangledown_{\boldsymbol{x}} \mathcal{L}(\boldsymbol{x},y^{target}))
  \label{eq:fgsmlosstarget}
\end{equation}

Kurakin et al.~\cite{Kurakin2016AdversarialEI} proposed a Basic iterative method (BIM) which is an iterative FGSM attack to achieve stronger attack effects through multiple rounds of small steps:
\begin{equation}
  \boldsymbol{x}^{adv}_{t+1}=Clip_{\boldsymbol{x},\epsilon}\{\boldsymbol{x}^{adv}_{t}+\alpha sign(\bigtriangledown_{\boldsymbol{x}} \mathcal{L}(\boldsymbol{x}^{adv}_{t},y))\}
  \label{eq:bimloss}
\end{equation}%
where $\boldsymbol{x}^{adv}_{t}$ represents the adversarial sample in the $t_{th}$ iteration, $\alpha$ represents the step size, and $Clip$ represents the truncation method.

Adversarial attacks can be broadly categorized based on the attack environment into white-box and black-box attacks. In white-box attacks, adversaries possess complete knowledge of the target model, while black-box attacks operate under the premise that the attackers lack direct access to the model's specific details.

Athalye et al.~\cite{Athalye2018ObfuscatedGG} introduced a gradient approximation method that adeptly circumvented the obfuscated gradients defense in a white-box context. They classified obfuscated gradient defenses into three types: broken gradients, random gradients, and vanishing/exploding gradients. Additionally, they proposed two distinctive attack strategies: Backward Pass Differentiable Approximation (BPDA) and Expectation Over Transformation (EOT).

A salient feature of adversarial samples is their inherent transferability across varying model architectures and parameterizations. Transferability paves the way for the feasibility of black-box attacks~\cite{Liu2016DelvingIT}, even facilitating their extension from digital settings to real-world scenarios~\cite{Thys2019FoolingAS}.


\subsection{Defense against adversarial attack}

For the deployment of intelligent models in real-world scenarios, it's imperative that models maintain accurate predictions even when confronted with adversarial attacks. Ilyas et al.~\cite{Ilyas2019AdversarialEA} referred to the model's ability to resist adversarial attacks as adversarial robustness.

Adversarial training~\cite{Goodfellow2014ExplainingAH}, a model-specific defense, has garnered considerable attention. It seeks to bolster a model's adversarial robustness by retraining on the training set with adversarial samples added. However, recent studies have highlighted its limitations. Training on large datasets like ImageNet, coupled with the generation of ample adversarial samples, can be resource-intensive~\cite{Wong2020FastIB}. Wong et al.~\cite{Wong2020OverfittingIA} pointed out the potential detriment of excessive adversarial training, while Schmidt et al.~\cite{Schmidt2018AdversariallyRG} observed that achieving pronounced adversarial robustness demands a significantly larger sample size compared to standard high-accuracy models. Furthermore, Tsipras et al.~\cite{Tsipras2018RobustnessMB} noted that model accuracy and adversarial robustness are negatively correlated. Such challenges impede the widespread adoption of adversarial training.

\begin{figure*}[t!]
  \centering
  \includegraphics[width=\linewidth]{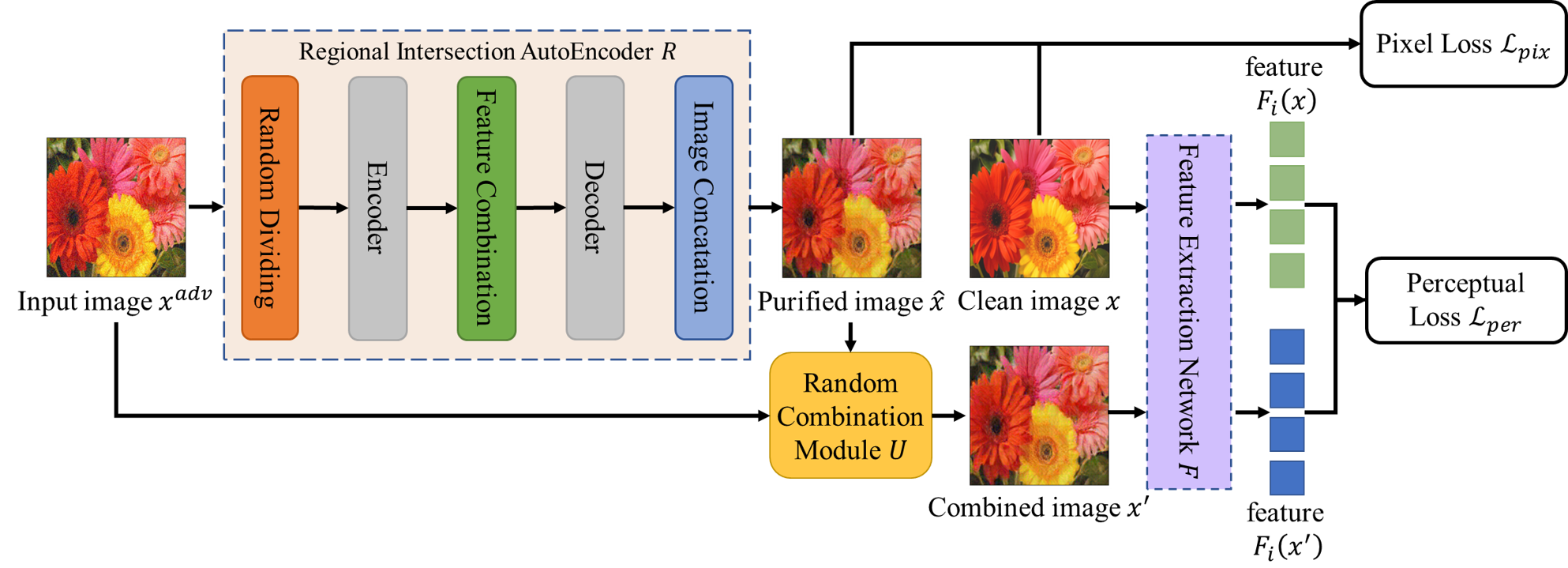}
  \caption{The framework of our adversarial purification method IMPure. We input the adversarial image into the image reconstruction network to get the purified image and calculate the pixel loss with the clean image stable low-level information. Then the adversarial image and the purified image are input to RCM to obtain the combined image. The feature extraction network extracts high-level features from combined and clean images and computes the loss to stabilize the semantic information.}
  \label{fig:OverviewAdvPuri}
\end{figure*}

Model-agnostic defenses primarily revolve around preprocessing the input image to neutralize adversarial perturbations. Pioneering efforts revealed that basic image transformations, such as random padding~\cite{Xie2017MitigatingAE} and JPEG compression~\cite{Dziugaite2016ASO}, could effectively thwart FGSM attacks. However, these rudimentary methods falter against sophisticated adversarial techniques and can also degrade image quality. Subsequent research shifted towards sophisticated approaches like denoising and reconstruction to restore clean images. For instance, Gu et al.~\cite{Gu2014TowardsDN} first proposed the use of denoising autoencoders to remove adversarial perturbations. Liao et al.~\cite{Liao2017DefenseAA} pointed out the disadvantages of pixel loss and introduced the High-Level Representation Guided Denoiser (HGD). Mustafa et al.~\cite{Mustafa2019ImageSA} combined wavelet denoising with super-resolution to enhance image quality, though its efficacy against potent strong adversarial attacks remains questionable. Jie et al.~\cite{Jia2018ComDefendAE} rolled out ComDefend, an end-to-end image compression technique, that sidesteps the need for a vast adversarial sample collection, albeit with limited defensive capabilities. Meng et al.~\cite{Meng2017MagNetAT} put forth MagNet, encompassing detectors and reformers to stave off adversarial attacks while preserving accuracy on standard inputs. Zhang et al.~\cite{Zhang2021DefenseAA} designed an image reconstruction network to eliminate the effects of adversarial perturbations. Nie et al.~\cite{Nie2022DiffusionMF} proposed that DiffPure purifies adversarial samples through forward and reverse processes of diffusion models, and residual perturbations may still attack successfully. Model-agnostic methods offer commendable versatility, yet several challenges persist. 

In this paper, we present a method aimed at addressing the challenges inherent in the aforementioned defense mechanisms. Firstly, the employment of perceptual loss often leads to the loss of fine-grained details, subsequently compromising model accuracy. To mitigate this, we introduce a dual constraint encompassing both pixel loss and perceptual loss, striving to retain essential image information. Secondly, fully eradicating adversarial perturbations remains challenging, and purifying images have potential security risks. In response, we devise RCM to emulate potential disturbance risks, fostering an active response from reconstruction models. Lastly, the defense of strong adversarial attacks is more challenging. We construct RIAE to resist the impact of strong adversarial perturbations through information masking and regional intersection reconstruction.

\section{Proposed Method}

\begin{figure*}[t]
  \centering
  \includegraphics[width=0.95\linewidth]{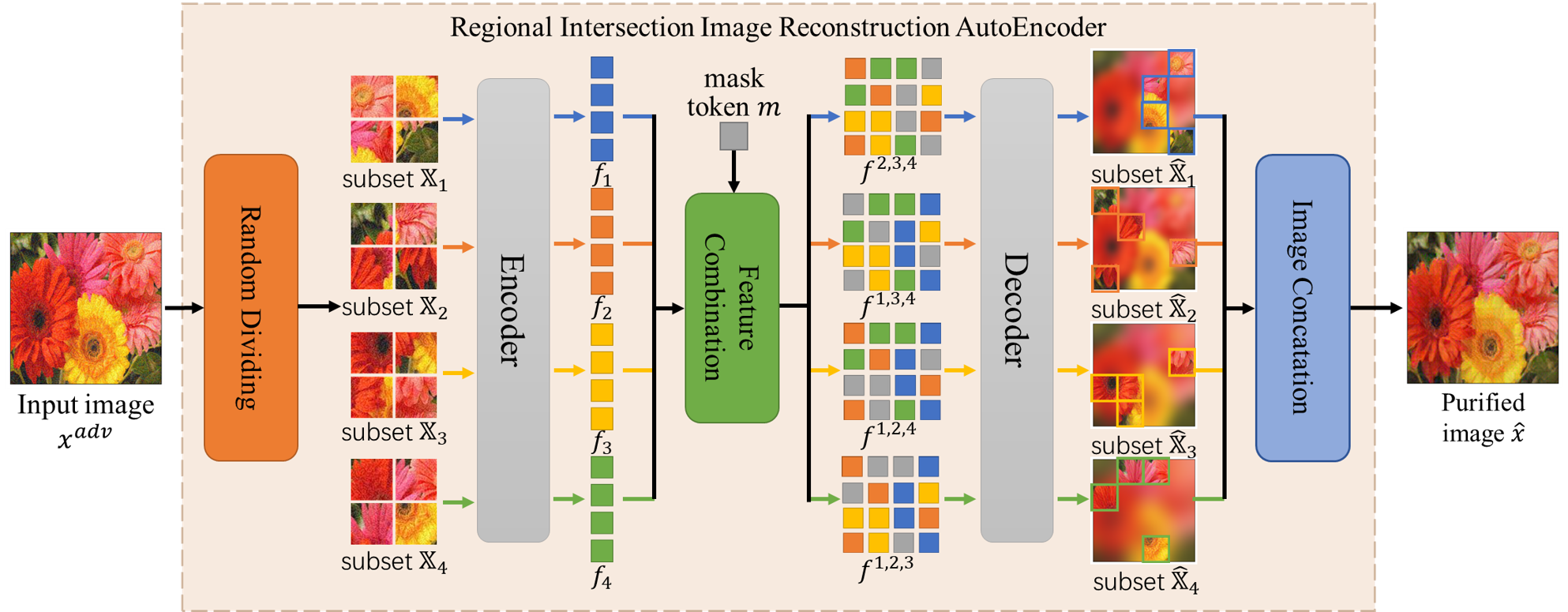}
  \caption{An overview of RIAE. The input image first is randomly divided into several subsets and input into the Encoder to output features. The features are combined and replaced with mask tokens to replace the parts that need to be masked, and then input into the Decoder respectively to output the purified subset. Finally, the patches in the subsets are combined to obtain the purified image.}
  \label{fig:ArchReconAE}
\end{figure*}

In this section, we first probe the impact of residual adversarial perturbations on the purified image and furnish a quantitative analysis linking perturbation scale to attack capability (Section \ref{sec:difflg}). Recognizing the pivotal role of curtailing residual perturbation scales in enhancing adversarial purification efficacy, we propose IMPure (Section \ref{sec:advpurimethod}). A schematic representation of our proposed defense model is depicted in Fig. \ref{fig:OverviewAdvPuri}. Given the difficulty of diminishing the aggregate perturbation scale of a purified image, our approach pivots on image patch processing. The perturbations in an image patch predominantly originate from the same-position patch and content-similar patch of the adversarial sample. To tackle perturbations from identical position patches, we put forth RIAE, visualized in Fig. \ref{fig:ArchReconAE}. Adversarial perturbations are mitigated by masking part of the image patches information. Then, the reconstruction model uses the information from the remaining image patches to reconstruct the masked image patches. The reconstruction process is executed in parallel on all patches to obtain an overall purified image. Addressing content-similar patches, RCM is introduced, bypassing intricate content computations. Operational during the training phase, RCM intentionally retains part of the adversarial perturbations on the purified image to simulate perturbations from content-similar patches. RIAE generates the purified image with stronger defense capabilities to resist the adverse effects of preservation perturbations on the target model. Conclusively, we design a joint constrained loss integrating pixel loss and perceptual loss, and explore a more ﬂexible perceptual loss design scheme. The joint loss is designed to encourage the model to achieve a virtuous cycle between low-level information and high-level representation.

\subsection{Theoretical Analysis}
\label{sec:difflg}

We begin with a brief overview of adversarial attack principles and the mechanisms of adversarial purification. Subsequently, we delve into the potential risks posed by residual adversarial perturbations on purified images to the target model. We establish a quantifiable correlation between perturbation scale and attack efficacy using a streamlined target model. Additionally, we discuss the influence of different residual conditions of perturbations on attack capability, leading us to assert the importance of constraining the scale of residual disturbances.

Adversarial attack methodologies are orchestrated with the primary intent of manipulating the target models to yield incorrect predictions. These techniques can generate small perturbations, which are added to clean images to obtain adversarial samples. It is difficult for the human eyes to detect the difference between clean images and adversarial samples, but adversarial samples can induce the target model to make wrong predictions. To formalize, consider an efficient target model $T$ and an input image $x$, yielding a prediction $y=T(\boldsymbol{x})$. An adversarial perturbation $\boldsymbol{\eta}$ is derived via an adversarial attack methodology and generates an adversarial sample $\boldsymbol{x}^{adv}=\boldsymbol{x}+\boldsymbol{\eta} $, subsequently causing the target model $T$ to produce incorrect predictions, i.e., $T(\boldsymbol{x}^{adv})\neq y$. 

Adversarial purification serves as a defensive mechanism against adversarial attacks, wherein the image is subjected to a “purification” process prior to being fed into the target model. An adversarial purification model $P$ receives an adversarial sample $x^{adv}$ and yields a purified sample $\hat{\boldsymbol{x}}=P(\boldsymbol{x}^{adv})$, ensuring that $T(\boldsymbol{x})=T(\hat{\boldsymbol{x}})$. 

Ideally, we hope to completely remove the adversarial perturbation $\boldsymbol{\eta}$, making the purified sample $\hat{\boldsymbol{x}}$ exactly the same as the clean image $\boldsymbol{x}$. Moreover, current research~\cite{Liao2017DefenseAA} shows that the remaining adversarial perturbations can still successfully attack, which inspired us to design adversarial purification methods to remove the adversarial perturbations and add defensive perturbations.

We formally discuss why simply reducing adversarial perturbations cannot achieve good defense results. We define a simple linear model $f(\boldsymbol{x})=\boldsymbol{\omega}^{T}\boldsymbol{x}$, a clean input $\boldsymbol{x}$ and an adversarial sample $\boldsymbol{x}^{adv}=\boldsymbol{x}+\boldsymbol{\eta} $, where $\boldsymbol{\omega}$ is a weight vector, $\boldsymbol{\omega}$ and $\boldsymbol{x}$ follow the standard normal distribution $N(0, 1)$, $\boldsymbol{\eta}$ is a small adversarial perturbation subject to the constraint ${\left\lVert \boldsymbol{\eta} \right\rVert}_{\infty } < \epsilon $, and $\epsilon$ is a small constant representing the perturbation threshold used to ensure the concealment of the adversarial sample. When the input of the linear model $f$ is an adversarial sample $\boldsymbol{x}^{adv}$, the output of model $f$ is
\begin{equation}
  \begin{aligned}
 f(\boldsymbol{x}^{adv})&=\boldsymbol{\omega}^{T}\boldsymbol{x}^{adv} \\
 &=\boldsymbol{\omega}^{T}\boldsymbol{x} + \boldsymbol{\omega}^{T}\boldsymbol{\eta }.
  \end{aligned}
  \label{eq:OutputByAdv}
\end{equation}%
The difference caused by the adversarial perturbation is
\begin{equation}
  \begin{aligned}
  \varDelta f &= f(\boldsymbol{x}^{adv}) - f(\boldsymbol{x}) \\
  &= \boldsymbol{\omega}^{T}\boldsymbol{\eta }.
  \end{aligned}
  \label{eq:DiffByAdv}
\end{equation}%
We maximize the difference by assigning $\boldsymbol{\eta }=\epsilon sign(\boldsymbol{\omega})$. At this time, the attack caused by the perturbation is the strongest, and the difference $\varDelta f = \epsilon {\left\lVert \boldsymbol{\omega}\right\rVert}_{1} $. If $\boldsymbol{\omega}$ has $n$ dimensions and the average magnitude of an element of the weight vector is $m$, then the expectation of the maximum difference $\mathbb{E}(\varDelta f)=\epsilon mn$. $n$ is usually very large in neural networks so even a small perturbation can cause a large difference in the prediction. Although many neural networks add nonlinearity through such activation functions ReLU, they still retain linearity as a whole, making it difficult for the network to resist adversarial perturbations.

Next, we discuss what happens after the adversarial purification method reduces adversarial perturbations. The previous discussion demonstrated that the attack capability of the adversarial sample varies linearly with the scale of the perturbation. Because the pixel value of the image is a discrete value of $[0, 255]$, if a perturbation threshold against perturbation is 16 unless we can completely remove the perturbation, even if the remaining perturbation is only 1, its attack capability remains $1/16$. Since the powerful attack capability against perturbations mainly comes from the high dimensionality of the weight vector, limited linear reduction still retains considerable attack capability. Therefore, simply reducing confrontational disturbances cannot achieve the reliable defense effects.

Since global reduction has a limited effect against disturbance, are other reduction methods effective? For example, compared with reducing the overall perturbation to half, would it be better to completely remove the purification in half of the area to achieve a better defensive effect? We define the scale of an adversarial perturbation as the sum of the absolute values of the perturbation and design two methods to reduce the adversarial perturbation: global reduction $R_g(\boldsymbol{\eta}, s)$ and local reduction $R_l(\boldsymbol{\eta} , s)$, where $s\in (0, 1)$ represents the multiple of the scale. The global method is to multiply the adversarial perturbation by $s$:
\begin{equation}
  R_g(\boldsymbol{\eta} , s) =s\boldsymbol{\eta} .
  \label{eq:GlobalReduce}
\end{equation}%
The local method is to retain $s$ of the area of the adversarial perturbation and assign 0 to the rest:
\begin{equation}
  R_l(\boldsymbol{\eta} , s) =\boldsymbol{v} \odot \boldsymbol{\eta},
  \label{eq:LocalReduce}
\end{equation}%
where $\boldsymbol{v}$ is a mask vector with the same shape as $\boldsymbol{\eta}$ in which $s n$ elements are 1 and the remaining elements are 0. $\odot $ specifies the element-wise multiplication. Two methods reduce adversarial perturbations to the same scale and then compare the attack capabilities of their processed adversarial samples. When the global method $R_g$ and the local method $R_l$ are used to reduce the adversarial perturbation $\boldsymbol{\eta }$ respectively, expectations of the maximum difference $\varDelta f_g$ and $\varDelta f_l$ are:
\begin{align}
  &\mathbb{E}(\varDelta f_g) = \mathbb{E}(\boldsymbol{\omega}^{T} s \boldsymbol{\eta }) = s\epsilon mn,
  \label{eq:DiffByGlobal} \\
  &\mathbb{E}(\varDelta f_l) = \mathbb{E}(\boldsymbol{\omega}^{T} (\boldsymbol{v} \odot \boldsymbol{\eta })) = s\epsilon mn.
  \label{eq:DiffByLocal}
\end{align}%
It is observed that two different reduction methods yield equivalent outcomes, with the attack capability against the perturbation varying linearly with scale. This shows that for a generative model with a limited ability to eliminate perturbations, the ability to defend against adversarial attacks is closely related to the scale of perturbations that the model can eliminate. Combined with the previous results, we consider that the complete elimination of adversarial perturbations is necessary for adversarial purification methods.

\subsection{Information Mask Purification}
\label{sec:advpurimethod}

In this section, we propose IMPure based on the theory in Section \ref{sec:difflg}. IMPure is architecturally composed of three principal components: RIAE, RCM, and a Feature Extraction Module. RIAE is responsible for transforming the input adversarial samples $\boldsymbol{x}^{adv}$ into purified samples $\hat{\boldsymbol{x}}$ without attack capabilities. The region intersection design we propose allows the network to reconstruct image patches without being interfered with by the same-position patches' adversarial perturbations. We propose RCM to encourage the generation of images with stronger defense capabilities. The adversarial samples $\boldsymbol{x}^{adv}$ and purified samples $\hat{\boldsymbol{x}}$ are spliced into a combined image $\boldsymbol{x}^{\prime}$ which are fed into the feature extraction network to obtain features for calculating the perceptual loss. To improve the effectiveness of the reconstruction network, we use joint constraints of pixel loss and feature loss to stabilize the underlying information and semantic information respectively. To this end, we extract high-level features of the purified image $\hat{\boldsymbol{x}}$ and the clean image ${\boldsymbol{x}}$ respectively through pre-trained feature extraction network for calculating the feature loss. The structure of IMPure is illustrated in Fig. \ref{fig:OverviewAdvPuri}.

\subsubsection{Regional Intersection AutoEncoder}
\label{sec:reconae}

\begin{figure}[htb]
  \centering
  \includegraphics[width=0.9\linewidth]{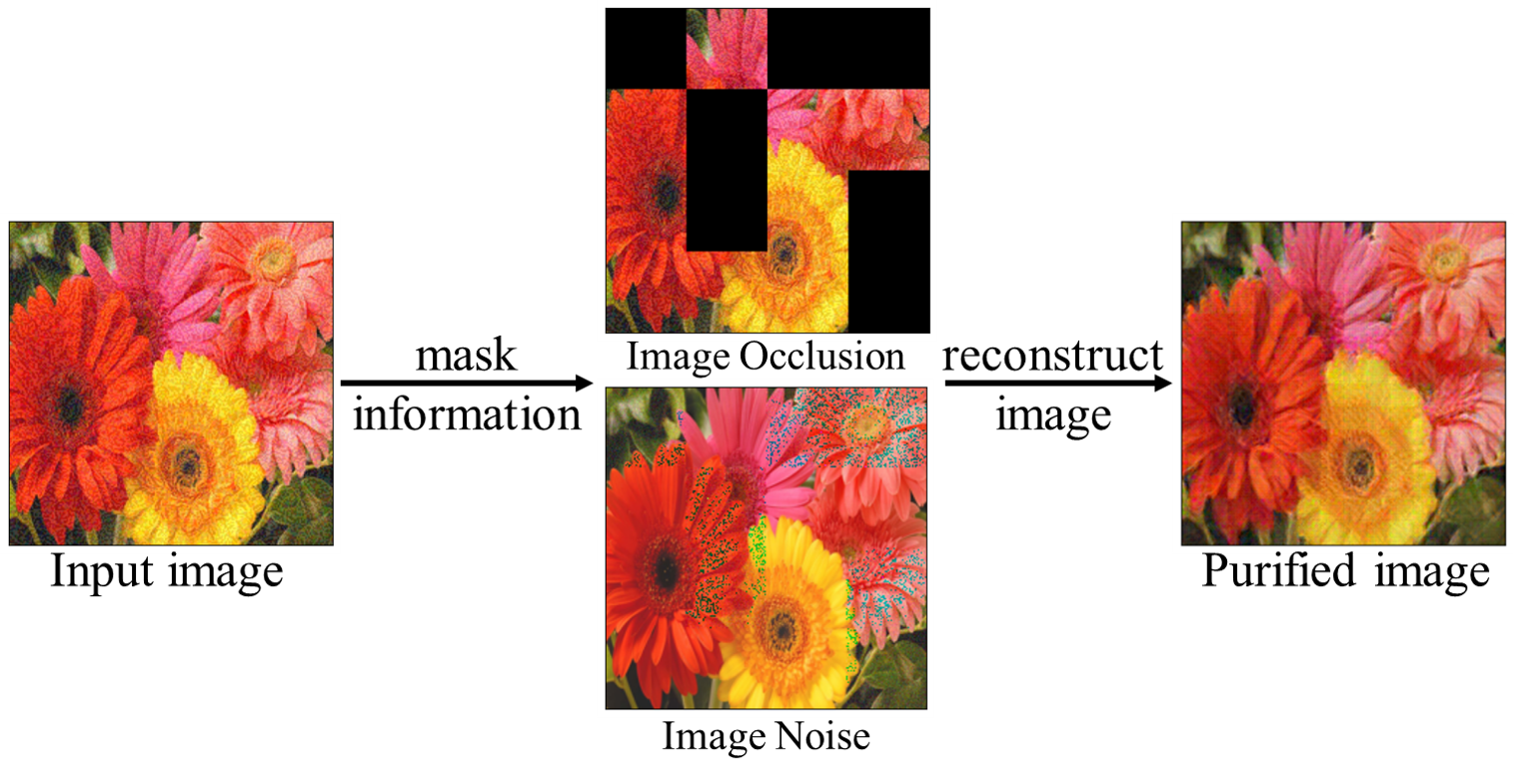}
  \caption{Information masking process: Get an input image, first use the information masking method on the specified area, and then reconstruct the masked image to obtain the purified image. Two information masking methods are illustrated respectively: Image Occlusion and Image Noise.}
  \label{fig:InformaskProcess}
\end{figure}

Given the discrete nature of image pixel values, limiting the overall perturbation scale of an image is difficult. Therefore we choose to reconstruct partial image patches at a time to simplify the task, by reconstructing in parallel on all patches to get the complete image. As shown in Fig. \ref{fig:InformaskProcess}, to further reduce the perturbation scale, we mask the information of the reconstructed patches. One straightforward method to remove perturbations in patches is to erase pixels, leading to a dual loss: both the adversarial perturbations and the underlying image information. In the experimental part, we also evaluated two other information masking methods, image noise and feature noise. We refer to this image reconstruction technique as ``Regional Intersection" and have developed a dedicated network RIAE for its implementation.

The architecture of the image reconstruction network is introduced below. In order to conveniently the operation of the image area, we build an autoencoder with reference to the Masked AutoEncoder (MAE)~\cite{He2021MaskedAA} structure. An overview of the autoencoder is shown in Fig. \ref{fig:ArchReconAE}. First, like Vision Transformer (ViT)~\cite{Dosovitskiy2020AnII}, an image $\boldsymbol{x} \in \mathbb{R}^{H\times W\times C}$ is divided into patches $\boldsymbol{x}_p \in \mathbb{R}^{N\times (P^2\cdot C)}$ and mapped into embedded patches $\boldsymbol{z} \in \mathbb{R}^{N\times D}$ through Patch embedding and Position embedding, where $(H, W)$ is the resolution of the image, $C$ is the number of the image channels,$(P, P)$ is the resolution of each image patch, $N=HW/P^2$ is the number of the patches, $D$ is the constant latent vector size for the transformer, Patch embedding is a trainable linear projection, and Position embedding is a fixed 2D sine-cosine encoding~\cite{Chen2021AnES} to simplify training, unlike ViT. 

Second, we treat the embedding patch as a set $\mathbb{X}=\{\boldsymbol{z}_{i,:}|i\in \{1,\cdots ,N\}\}$ and divide it into several subsets $\{\mathbb{X}_i\|i\in \{1,\cdots,S\}\}$, where $S$ is the number of subsets we specify. Randomly shuffle the order of patches and divide them into subsets evenly. The subsets are equivalent to uniform sampling of the image, and there are no repeated patches between subsets. Then just like standard ViT, we process these subsets with a series of transformer blocks to obtain the feature set $\mathbb{F}=\{\boldsymbol{f}_i\in \mathbb{R}^{L\times D}\}$, where $L=N/S$ is the number of patches in the subset. Each subset is encoded separately, there is no data interaction between subsets, and adversarial perturbations in different subsets will not affect each other.

Third, for the decoder, we occlude a subset and input the other subsets to the decoder to reconstruct the mask patches. We perform this step for each subset and finally combine all reconstructed parts together to obtain the complete reconstructed image. In the specific implementation, we combine all subsets into complete features according to their original positions and replace the features of the subset to be occluded with a mask token $\boldsymbol{m}\in \mathbb{R}^{D}$, then add position embedding and use a series of transformer blocks to complete the reconstruction of the subset set $\{\hat{\mathbb{X}}_i\|i\in \{1,\cdots, S\}\}$. Finally, we combine the patches in the subset set by location to obtain the purified image $\hat{\boldsymbol{x}}$.

\subsubsection{Random Combination Module}
\label{sec:combin}

Through the utilization of the aforementioned RIAE, we have procured purified images capable of counteracting adversarial attacks.  While the autoencoder remains ostensibly impervious to the adversarial perturbations within the corresponding regions, it's imperative to acknowledge that in realistic imagery, the presence of regions with analogous features is not uncommon, and the adversarial perturbations within such regions often exhibit similarity. This congruency in perturbations potentially renders the autoencoder susceptible to adversarial influence. To enhance defensive capabilities and mitigate this susceptibility, we introduce RCM. This module is meticulously designed to retain adversarial perturbations within random regions during the training phase intentionally. Such strategic retention is aimed at incentivizing the image reconstruction network to synthesize purified images endowed with fortified defensive attributes. This augmentation in the training paradigm serves to bolster the resilience of the autoencoder against adversarial perturbations, even those exhibiting subtle similarities within different regions, thereby refining the robustness of the overall model.

Our combination strategy is to randomly generate a mask matrix $\boldsymbol{u}\in \mathbb{R}^{H\times W}$ with the same resolution as the image, with each pixel
having a greyscale value between 0.0 and 1.0, and obtain the combined image $\boldsymbol{x}^{\prime}$ through calculation:
\begin{equation}
  \boldsymbol{x}^{\prime} = \hat{\boldsymbol{x}} \odot \boldsymbol{u} + \boldsymbol{x} \odot (1-\boldsymbol{u}).
  \label{eq:RandomCombine}
\end{equation}%
In order to echo the transformer structure used by the image reconstruction network, the mask matrix also takes the same patch size to simulate the situation where the adversarial perturbations of some patches are not cleared.

\subsubsection{Loss function}
\label{sec:perpixloss}

We utilize a combination of reconstruction loss and perceptual loss to train our image reconstruction network $R$. $\theta$ represent the parameters of our network. Our network receives an input image $\boldsymbol{x}^{adv}$ and outputs a purified image $\hat{\boldsymbol{x}}$:
\begin{equation}
  \hat{\boldsymbol{x}}=R({\boldsymbol{x}^{adv}},\theta).
  \label{eq:NetworkInOut}
\end{equation}%
The pixel-level loss as the reconstruction loss is defined as:
\begin{equation}
  \mathcal{L}_{pix}(\hat{\boldsymbol{x}}, \boldsymbol{x}, \theta )=\frac{1}{H W }{\left\lVert \hat{\boldsymbol{x}}-\boldsymbol{x}\right\rVert}_1,
  \label{eq:PixLoss}
\end{equation}%
where $\boldsymbol{x} $ is a clean image paired with $\boldsymbol{x}^{adv}$. We utilize $L_1$ loss which has been demonstrated to better than $L_2$ loss for image restoration~\cite{Zhao2017LossFF}.

Adversarial images and clean images are often very similar at the pixel level but have great differences in the feature representation of the target model, which leads to different predictions by the target model. In order to make the purified image consistent with the clean image in terms of feature representation, we use perceptual loss for constraints. Specifically, we record a convolutional network used for feature extraction as $F $, and ${F}_i(\boldsymbol{x})$ is the feature map output by the image $\boldsymbol{x}$ at the $i$-th layer of the network $F $. The perceptual loss is defined as the difference in feature representation between the combined image $\boldsymbol{x}^{\prime}$ output by the image combination module and the clean image $\boldsymbol{x}$: 
\begin{equation}
  \begin{aligned}
  \mathcal{L}_{per\_i}(\boldsymbol{x}^{\prime}, \boldsymbol{x}, \theta )&=D({F}_i(\boldsymbol{x}), {F}_i(\boldsymbol{x}^{\prime})) \\
  &=\frac{1}{H_{{F}_i} W_{{F}_i} } {\left\lVert {F}_i(\boldsymbol{x}) - {F}_i(\boldsymbol{x}^{\prime})\right\rVert}_1,
  \label{eq:PerLoss}
  \end{aligned}
\end{equation}%
where $D$ is the distance function used to measure feature differences. In our method, the $L_1$ norm is used as the distance function, and $(H_{{F}_i}, W_{{F}_i})$ is the resolution of the feature map. 

The key to the success of perceptual loss is the structure of the feature extraction network~\cite{Liu2021GenericPL}. The model's feature representation usually has certain characteristics, hence it may be effective to choose the corresponding model as the feature extraction network. However, single-model selection can diminish the generalization of the defense method, while the overhead of multi-model integrated training remains substantial. Consequently, we elect to employ the 19-layer VGG network~\cite{Simonyan2014VeryDC} for feature extraction. First, the VGG network has proven its superiority in tasks using perceptual loss in other fields. Second, the convolutional structure of the VGG network is widely used by many CNNs. Finally, the network's depth plays a pivotal role in image feature extraction. To capture the correlation between multi-layer statistics extracted by multi-layer CNN, we integrate multi-layer feature maps in the perceptual loss. The perceptual loss is defined as:
\begin{equation}
  \mathcal{L}_{per}(\boldsymbol{x}^{\prime}, \boldsymbol{x}, \theta )=\sum_{K} \lambda _{i} \mathcal{L}_{per\_i}(\boldsymbol{x}^{\prime}, \boldsymbol{x}, \theta ),
  \label{eq:VGGPerLoss}
\end{equation}%
where $K$ represents the set of layers of VGG feature maps involved in the operation. $\lambda _{i}$ represents the weight of the loss. VGG19 contains 5 similar modules, each module contains convolution layers, pooling layers, and nonlinear activation layers. We select the feature map output after the ReLU activation layer in each module to participate in the calculation of perceptual loss. 

The overall loss we use to train our netword is defined as:
\begin{equation}
  \begin{aligned}
  \mathcal{L}_{overall}(\boldsymbol{x}^{adv}, \boldsymbol{x}, \theta )=&\mathcal{L}_{pix}(\hat{\boldsymbol{x}}, \boldsymbol{x}, \theta ) + \\
  &\sum_{K} \lambda _{i} \mathcal{L}_{per\_i}(\boldsymbol{x}^{\prime}, \boldsymbol{x}, \theta ),
  \end{aligned}
  \label{eq:JointLoss}
\end{equation}%
where we obtain the combination of losses through experiments select the feature maps of the 4-th and 5-th modules and the previous layer output of the softmax layer. We set $\lambda _{4}=30$, $\lambda _{5}=10$ and $\lambda _{pre\_soft}=5$.

\section{Experiments}
\label{sec:exper}

In this section, we illustrate the experiments and use large-scale results to prove the superiority of our method.

\subsection{Experimental Settngs}

{\bf Target models}: The target model architectures we choose are Inception V3 (IncV3)~\cite{Szegedy2016xceptionv3}, Inception-ResNet-v2 (IncRes)~\cite{Szegedy2017inceptionv4} and ResNet-v2-101 (ResNet)~\cite{He2016IdentityMI}, and the task is image classification. All models are pre-trained on ImageNet dataset, training settings and model weights can be found on this page\footnote{https://pytorch.org/vision/stable/models.html\#classification}. We do not make any modifications to the structure and weights of the model, and the testing of the model will be conducted according to the original settings.

{\bf Attack methods}: We select a series of adversarial attack methods to generate adversarial examples, including FGSM~\cite{Goodfellow2014ExplainingAH}, C\&W (CW)~\cite{Carlini2016TowardsET}, Deepfool (Df)~\cite{MoosaviDezfooli2015DeepFoolAS}, PGD~\cite{Madry2017TowardsDL}, MIFGSM (MI)~\cite{Dong2017BoostingAA}, DIFGSM (DI)~\cite{Xie2018ImprovingTO}, MDFGSM (MD)~\cite{Xie2018ImprovingTO}, APGD-ce (A-ce)~\cite{Croce2020ReliableEO} and APGD-dlr (A-dlr)~\cite{Croce2020ReliableEO}. The code implementation of the first five methods refers to the CleverHans library\footnote{https://github.com/cleverhans-lab/cleverhans}~\cite{papernot2018cleverhans}, and the remaining four come from the Torchattacks library\footnote{https://github.com/Harry24k/adversarial-attacks-pytorch}~\cite{kim2020torchattacks}. For methods that support targeted attacks and non-targeted attacks, we uniformly choose non-targeted attacks. To generate strong adversarial samples, the perturbation threshold $\epsilon $ is set to $16/255$, and other parameters of the attack method will use the default settings of the source code author.

{\bf Dataset}: We train and evaluate our defense method on the ImageNet~\cite{Deng2009ImageNetAL} dataset with top-1 classification accuracy, which is widely used in adversarial attack~\cite{Dong2017BoostingAA, Xie2018ImprovingTO, Croce2020ReliableEO} and defense~\cite{Xie2017MitigatingAE, Prakash2018DeflectingAA, Jia2018ComDefendAE} domains. For each target model, we only select images that the model can correctly classify, and modify the resolution of the image according to the input requirements of the model. We randomly select 10 images from each category in the Imagenet training set and then use all adversarial attack methods to generate adversarial samples as the training set for our defense method. We randomly select 5 images from each category in the Imagenet verification set to generate adversarial samples as the verification set. Use the same method to select different images to generate adversarial samples as the test set.

{\bf Implementation details}: The structure and parameter settings of the transformer part of Regional Intersection AutoEncoder refer to MFFAE~\cite{Liu2023ImprovingPM}. The patch size is set to $16\times 16$. For the transformer block of the encoder part, the embedding dimension is set to 768, the depth is set to 12, and the number of multi-nodes inside is set to 12. For the transformer block of the decoder part, the embedding dimension is set to 512, the depth is set to 8, and the number of multi-nodes inside is set to 16. Feature extraction network selection VGG19. We train our model using AdamW~\cite{Loshchilov2017DecoupledWD} with $\beta _1=0.9$, $\beta _2=0.95$. The learning rate is initially set to $1e^{-4}$ and use CosineAnnealing~\cite{Loshchilov2016SGDRSG} with 100 epochs. The training procedure is described in the Algorithm. \ref{alg:riadvpuritrainalg}.

\begin{algorithm}[htb]
  \caption{Regional Intersection Adversarial Purification Method Training Algorithm.}
  \label{alg:riadvpuritrainalg}
  \renewcommand{\algorithmicrequire}{\textbf{Input:}}
  \renewcommand{\algorithmicensure}{\textbf{Output:}}
  \begin{algorithmic}[1]
  \REQUIRE Adversarial image $\boldsymbol{x}^{adv}$, clean image $\boldsymbol{x}$, regional intersection autoencoder $R $ parameterized by $\theta $, random combine module $U $, feature extraction network $F $ with map layers $K$, learning rate $l_r$ and number of training epochs of $T$. 
  \STATE Initialize $\theta $ with random values;
  \FOR {$t \leftarrow 0$ to $T$} 
  \STATE $\hat{\boldsymbol{x}} \leftarrow R (\boldsymbol{x}^{adv}, \theta)$;
  \STATE $\boldsymbol{x}^{\prime} \leftarrow U (\hat{\boldsymbol{x}}, \boldsymbol{x}^{adv})$;
  \STATE Calculate $\mathcal{L}_{pix}$, $\mathcal{L}_{per}$ and $\mathcal{L}_{overall}$ using Eq. \ref{eq:PixLoss}, Eq. \ref{eq:VGGPerLoss} and Eq. \ref{eq:JointLoss}, respectively;
  \STATE $\theta \leftarrow \theta - l_r{\bigtriangledown}_{\theta}(\mathcal{L}_{overall}(\boldsymbol{x}^{adv}, \boldsymbol{x}, \theta ))$;
  \ENDFOR .
  \end{algorithmic}
\end{algorithm}

\begin{table*}[!htb]
  \centering
  \setlength{\belowcaptionskip}{0.2cm}
  \caption{Top-1 classification accuracy of the three target models on the test set without defense and with defense in a black-box setting.}
  \label{tab:blackboxeval}
  \setlength{\tabcolsep}{1mm}{
  \begin{tabular}{lccccccccccc} 
  \hline
  \multicolumn{2}{c}{Methods}                      & Clean          & FGSM~\cite{Goodfellow2014ExplainingAH}           & CW~\cite{Carlini2016TowardsET}             & Df~\cite{MoosaviDezfooli2015DeepFoolAS}       & PGD~\cite{Madry2017TowardsDL}            & MI~\cite{Dong2017BoostingAA}         & DI~\cite{Xie2018ImprovingTO}         & MD~\cite{Xie2018ImprovingTO}         & A-ce~\cite{Croce2020ReliableEO}        & A-dlr~\cite{Croce2020ReliableEO}        \\ 
  \hline
  \multirow{2}{*}{ResNet~\cite{He2016IdentityMI}}       & No defense & 100.00 & 53.70 & 1.20  & 0.12     & 0.74  & 0.42   & 2.24   & 2.26   & 0.56    & 0.36      \\
                                       & Defense    & 87.54  & 74.42 & 83.96 & 83.74    & 75.70 & 66.04  & 68.44  & 57.52  & 74.38   & 74.76     \\ 
  \hline
  \multirow{2}{*}{IncV3~\cite{Szegedy2016xceptionv3}}        & No defense & 100.00 & 26.40 & 0.22  & 0.00     & 0.04  & 0.06   & 0.16   & 0.08   & 0.08    & 0.84      \\
                                       & Defense    & 87.30  & 66.02 & 84.28 & 78.18    & 75.00 & 64.18  & 70.04  & 58.52  & 74.94   & 73.78     \\ 
  \hline
  \multirow{2}{*}{IncRes~\cite{Szegedy2017inceptionv4}} & No defense & 100.00 & 48.32 & 2.24  & 0.08     & 1.34  & 1.26   & 2.84   & 3.04   & 1.52    & 1.78      \\
                                       & Defense    & 85.86  & 70.88 & 82.40 & 80.34    & 76.44 & 66.52  & 71.26  & 60.74  & 75.20   & 74.38     \\
  \hline
  \end{tabular}}
  \end{table*}

\begin{table*}[!htb]
  \centering
  \setlength{\belowcaptionskip}{0.2cm}
  \caption{Top-1 classification accuracy of IncV3 against different defense methods on the test set in a black-box setting.}
  \label{tab:blackboxcompeval}
  \setlength{\tabcolsep}{1.5mm}{
  \begin{tabular}{lcccccccccc} 
    \hline
    Method    & Clean          & FGSM~\cite{Goodfellow2014ExplainingAH}           & CW~\cite{Carlini2016TowardsET}             & Df~\cite{MoosaviDezfooli2015DeepFoolAS}       & PGD~\cite{Madry2017TowardsDL}            & MI~\cite{Dong2017BoostingAA}         & DI~\cite{Xie2018ImprovingTO}         & MD~\cite{Xie2018ImprovingTO}         & A-ce~\cite{Croce2020ReliableEO}        & A-dlr~\cite{Croce2020ReliableEO}        \\ 
    \hline
    Attack    & 100.00         & 26.40          & 0.22           & 0.00           & 0.04           & 0.06           & 0.16           & 0.08           & 0.08           & 0.84            \\
    Random~\cite{Xie2017MitigatingAE}    & \textbf{95.56} & 36.18          & 37.86          & 25.30          & 3.70           & 1.34           & 0.34           & 0.42           & 4.52           & 15.30           \\
    PD+WD~\cite{Prakash2018DeflectingAA}     & 80.34          & 36.78          & 58.22          & 37.64          & 15.26          & 3.00           & 7.54           & 1.54           & 12.48          & 24.56           \\
    WD+SR~\cite{Mustafa2019ImageSA}      & 84.84          & 48.34          & 78.50          & 69.72          & 42.40          & 29.40          & 32.84          & 20.44          & 40.30          & 49.70           \\
    ComDefend~\cite{Jia2018ComDefendAE}  & 89.44          & 36.92          & 83.40          & 54.98          & 38.78          & 10.10          & 17.30          & 2.84           & 45.74          & 51.30           \\
    Recon~\cite{Zhang2021DefenseAA}     & 94.02  & 64.58 & 52.48 & 42.26    & 60.84 & 55.36  & 43.82  & 43.24  & 65.16   & 67.66     \\
    DIR~\cite{Zhou2023EliminatingAN}       & 64.02  & 53.56 & 63.02 & 61.58    & 58.92 & 52.32  & 55.88  & 48.96  & 57.40   & 57.16     \\
    Ours      & 87.30          & \textbf{66.02} & \textbf{84.28} & \textbf{78.18} & \textbf{75.00} & \textbf{64.18} & \textbf{70.04} & \textbf{58.52} & \textbf{74.94} & \textbf{73.78}  \\
  \hline
  \end{tabular}}
  \end{table*}

\subsection{Black-box evaluation}

We assess the efficacy of our proposed method in a black-box setting. The black-box scenario where it is presumed that the attacker has access to the structure and parameters of the target network, yet remains unaware of the defense mechanism. Table \ref{tab:blackboxeval} delineates the top-1 classification accuracy of three target models on the test dataset, both in the absence and presence of defense. Each attack methodology is specified under the ``Methods" row, with ``Clean" denoting the original unaltered sample. The ``No defense" row reflects the classification accuracy of the target model when subjected to adversarial samples; a lower value here signifies a more potent attack methodology. The ``Defense" column represents the classification accuracy of the purified images, with higher values indicating enhanced adversarial robustness.

The adversarial robustness of the target model can be ascertained through metrics presented in the ``No defense" row. IncRes exhibits the best adversarial robustness in the face of adversarial attacks, while IncV3 performs the worst. Nevertheless, the substantial dip in classification accuracy witnessed across all three target models underscores the imperativeness of defensive strategies.

Post-defense application, a marginal decline in the ``Clean" metrics across all three target models is observed. This emanates from the information masking our method enacts during image reconstruction, inevitably leading to some loss of image information. In juxtaposition with the ``No defense" scenario, a marked improvement is discernible across various attack methodologies once the defense is activated. Overall, IncRes continues to exhibit the most robust adversarial resilience.

Different attack samples perform differently under defense strategies. Attacks via C\&W and DeepFool are relatively easier to thwart as these techniques prioritize stealth and employ L2 norm constraints to mitigate perturbations, thereby moderating their attack potency to just the threshold of effectiveness, rendering them more easily defensible. In stark contrast, warding off attacks from MIFGSM, DIFGSM, and MDFGSM proves more challenging due to their emphasis on attack migration. APGD, an enhanced variant of PGD, presents an even tougher defense challenge. It's noteworthy to mention that given Inception's dismal accuracy against FGSM attack samples, the defense metrics too, are correspondingly low.

\begin{figure*}[!htb]
  \centering
  \includegraphics[width=\linewidth]{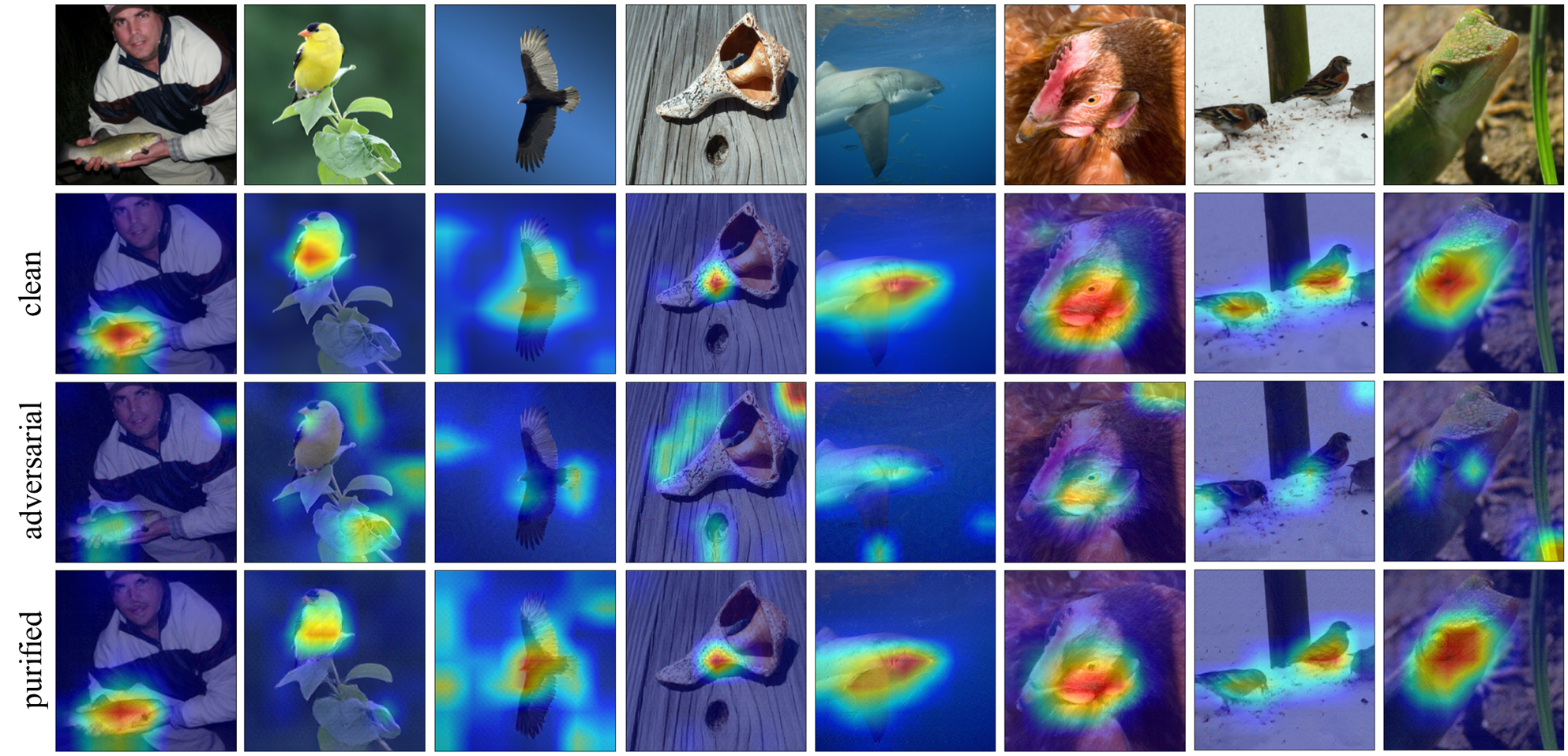}
  \caption{Class activation maps of GradCAM on resnet101 for clean images, adversarial examples and sanitized images.}
  \label{fig:HotAtt}
\end{figure*}

\subsection{Comparison with previous state-of-the-art methods}

In order to prove the effectiveness of our proposed defense model, we selected four model-agnostic methods for comparison, including random resizing and padding (Random)~\footnote{https://github.com/cihangxie/NIPS2017\_adv\_challenge\_defense}\cite{Xie2017MitigatingAE}, pixel deflection(PD+WD)\footnote{https://github.com/iamaaditya/pixel-deflection}~\cite{Prakash2018DeflectingAA}, wavelet denoising \& super-resolution(WD+SR)\footnote{https://github.com/aamir-mustafa/super-resolution-adversarial-defense}~\cite{Mustafa2019ImageSA}, ComDefend\footnote{https://github.com/jiaxiaojunQAQ/Comdefend}~\cite{Jia2018ComDefendAE}, Recon\footnote{https://github.com/ZOMIN28/Reconstructing-Images}~\cite{Zhang2021DefenseAA} and DIR\footnote{https://github.com/dwDavidxd/DIR}~\cite{Zhou2023EliminatingAN}. Among them, Random and PD are static defense strategies, and WD+SR, ComDefend, Recon and DIR include deep models. We used the weights provided in the WD+SR project. ComDefend, Recon and DIR did not provide appropriate weights, so we trained the weights ourselves to participate in the test set in strict accordance with the project requirements. All testing procedures are carried out strictly in accordance with the requirements in their projects. We compared the classification accuracy of various adversarial examples of IncV3 on the test set. The higher the classification accuracy, the more effective the defense against this type of attack is.

The ``Clean" metric illuminates the capability of the defense technique to retain the fundamental information within the image. Random emerges as the top performer, attributing to the fact that resizing and padding operations preserve most of the image information, and are frequently employed in diverse data augmentation techniques. On the other hand, Recon has obvious advantages in maintaining image information based on the Unet-like model, rendering its performance slightly inferior to Random. ComDefend executes compression operations, consequently sacrificing some level of detailed information. Similarly, our approach induces some information loss due to information masking, aligning its performance closely with that of ComDefend. Given that CNN lacks scale invariance, the target model is difficult to classify the super-resolution images generated by WD+SR. The PD method, which involves pixel swapping, disrupts local information, exerting a significant impact on the classification task. The image erasure used by DIR loses a lot of information when facing high-resolution images, which greatly affects the quality of the purified image and limits the defense performance.

The ensuing metrics unveil the defense method's ability to resist specific adversarial attacks. Our defense method achieves the best defensive performance against all attacks, achieving conspicuous advantages, especially in the three strong adversarial attack methods of MIFGSM, DIFGSM, and MDFGSM. Static methods, Random and PD, exhibit a semblance of defensive potential against adversarial samples characterized by weaker attack capabilities, however, their efficacy dwindles against robust attack modalities. Both WD+SR and ComDefend provide defense against all attacks, albeit their performance is circumscribed by their exclusive reliance on pixel loss.

\subsection{Effect of the adversarial perturbation on the feature map}

Adversarial perturbations are usually invisible to the human eye, but when added to a clean image, they cause very noticeable changes in the intermediate feature maps of the target model. This perturbation usually amplifies as the network propagates layer by layer, eventually causing the target model to output incorrect predictions. By visualizing and comparing the feature maps of clean images, adversarial samples, and purified images in the target model, the defense effect of the adversarial purified model can be intuitively demonstrated. As shown in Fig. \ref{fig:HotAtt}, we used GradCAM~\cite{Selvaraju2016GradCAMVE} to visualize the class activation map of the image on resnet101. Adversarial examples are generated by the PGD method. We can clearly see that the areas of focus for the clean image target model are closely related to the targets in the image. For adversarial samples, there is no obvious correlation between the area of interest and the target. The clean images generated by our adversarial cleansing method are basically the same as the clean images on the class activation map.



\section{Ablation experiment}

To further optimize the network structure, we try various combinations and optimization strategies to construct our methods and compare their performance. In order to simplify the problem, we use ResNet as the target model, the training epoch is set to 30, and other experimental settings remain consistent with the previous section.

\subsection{Effects of difference information mask methods}

\begin{figure}[htb]
  \centering
  \includegraphics[width=1.0\linewidth]{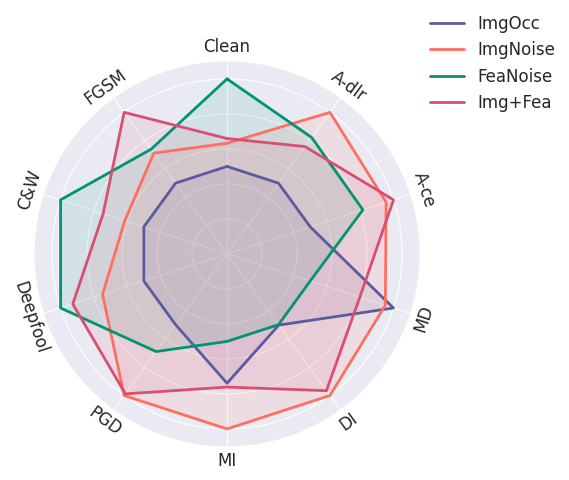}
  \caption{Performance of defense models against various adversarial attack methods under different information mask methods.}
  \label{fig:informask}
\end{figure}

In our proposed framework, the information masking technique is pivotal. An optimal masking method ought to retain the innate information of the image to the maximum extent while neutralizing a majority of the adversarial perturbations. We have instituted three distinct information masking methods: Image Occlusion (ImgOcc), Image Noise (ImgNoise), and Feature Noise (FeaNoise). ImgOcc epitomizes the total elimination of image information, standing as the most radical among the trio, as it retains neither the adversarial perturbation nor the original information. ImgNoise involves the injection of Gaussian noise with random intensity into the image. Considering that the pixel value that resists disturbance is small, a noise of suitable intensity can resist most perturbations, preserving the original information to a certain degree, thereby facilitating image reconstruction. FeaNoise operates on the features yielded by the RIAE encoder, introducing Gaussian noise of random intensity to these features. As depicted in Fig. \ref{fig:informask}, we conduct an evaluation to gauge the efficacy of the three information masking methods alongside the amalgam of ImgNoise and FeaNoise (Img+Fea). To render a lucid comparative insight via a singular graph, we normalized the results.

The Clean indicator succinctly illustrates the image reconstruction capability of the model and echoes the information loss induced by the information masking method. Our observations reveal the most subpar performance with ImgOcc, given its failure to transmit any information. ImgNoise exhibits a marginally better performance as the Gaussian noise retains some low-frequency information. FeaNoise outshone owing to its model's proficiency in learning anti-noise features, thus being minimally impacted. Reflecting upon the defense performance, ImgOcc is markedly inferior across most attack methods. However, it matched FeaNoise on DIFGSM, outperformed FeaNoise on MIFGSM, and outstripped both ImgNoise and FeaNoise on MDFGSM. This underscores the superior perturbation resistance of ImgOcc over ImgNoise and FeaNoise. In the battles of FGSM, CW, and DeepFool, FeaNoise trumps ImgNoise, although it falls significantly behind ImgNoise in other attack methodologies, revealing a better disturbance resistance of ImgNoise over FeaNoise. Finally, the composite method of Img+Fea manifests the pinnacle of overall performance, demonstrating that both masking methods are individually resistant to unique perturbations. ImgNoise+FeaNoise is thereby employed in our defense method.


\subsection{Effects of difference perceptual losses}

\begin{figure}[htb]
  \centering
  \includegraphics[width=1.0\linewidth]{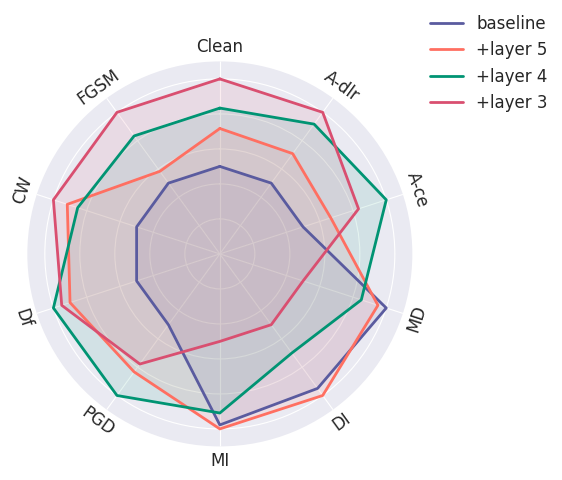}
  \caption{Performance of defense models against various adversarial attack methods under different perceptual loss settings.}
  \label{fig:Perloss}
\end{figure}

We evaluate the resulting performance of various loss combinations on lightweight experimental designs to obtain optimal losses. The output of the RELU layer in each convolution module of VGG19 will participate in the calculation of perceptual loss. We mark the layer in which the i-th module participates in the operation as layer i. As shown in Fig. \ref{fig:Perloss}, we set the initial perceptual loss to the previous layer output of the softmax layer of VGG19, use the pixel loss plus the perceptual loss as the benchmark, and then add one layer of feature maps to the perceptual loss each time and observe the experimental results.

On the three indicators of Clean, FGSM, and APGD-dlr, the performance gets better as the number of feature maps increases. The shallower the feature map, the lower the level of information it represents. This low-level information is obviously helpful for the reconstruction of clean images. At the same time, the experimental results also show that these shallow features are also of positive significance for weak adversarial attacks. In contrast, on MIFGSM, DIFGSM, and MDFGSM, the performance decreases as the number of feature maps increases. These three adversarial attacks are the three with the worst defense effects among the various defense methods currently available. For such adversarial samples with robust attack capabilities, shallow features can no longer have a positive effect. Based on various indicators, our defense method finally chose to add layer5+layer4 to the perceptual loss.

\subsection{Effects of difference dividing methods}

\begin{figure}[htb]
  \centering
  \includegraphics[width=\linewidth]{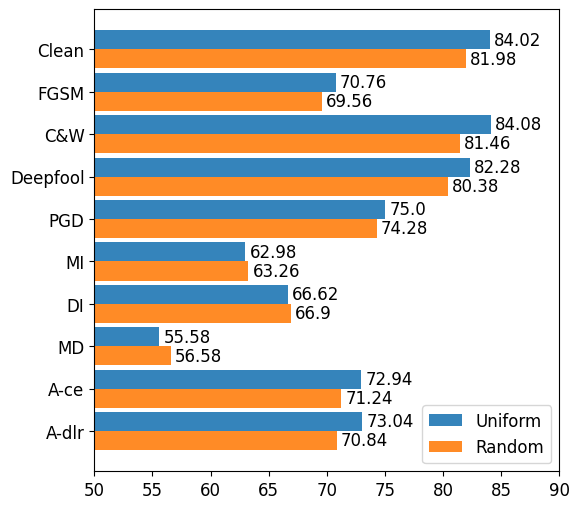}
  \caption{Top-1 classification accuracy of defense models against various adversarial attack methods for two dividing methods.}
  \label{fig:Sampling}
\end{figure}

The image is partially occluded and then reconstructed after the input area Regional Intersection image reconstruction AutoEncoder. In the training phase, in order to prevent overfitting, we randomly select some areas for occlusion. During the testing phase, we experimentally determined which method of selecting occlusion areas would achieve the best performance. The two dividing methods are: a uniform method is used to select the area, and the same random selection as in the training stage. We show in Fig. \ref{fig:Sampling} the performance of the defense model in resisting various adversarial attack methods using two methods. In order to visually demonstrate the performance gap between different methods, we normalized the results. The uniform method has obvious advantages over the random method in most indicators. The random method's defense performance against MIFGSM, DIFGSM, and MDFGSM is slightly better than the uniform method. This may be due to the fact that the adversarial perturbations generated by these robust adversarial attack methods can also take effect in adjacent areas, making it difficult for uniform methods to resist. We consider the gap in comprehensive performance between the two methods and choose to use the uniform dividing method during testing.

\subsection{Effects of defense against agnostic attacks}

Agnostic attacks denote adversarial assault mechanisms that remain external to the model training regimen. Given the incessant evolution of adversarial attack methodologies, fortifying models against agnostic attacks has emerged as a crucial endeavor. We conduct experiments to appraise the robustness of our defense strategy against gradient-based agnostic onslaughts. Early gradient-based attack variants such as FGSM, PGD, and MIFGSM are integrated into the training dataset, while DIFGSM, DMFGSM, and APGD are employed as agnostic attack exemplars to evaluate the defense efficacy of the model. The outcomes, delineated in Table \ref{tab:agnosticattack}, affirm that our defense paradigm continues to exhibit commendable resilience when confronted with agnostic assaults.

\begin{table}[htb]
  \centering
  \setlength{\belowcaptionskip}{0.2cm}
  \caption{Top-1 classification accuracy of ResNet against against agnostic attacks on the test set in a black-box setting.}
  \label{tab:agnosticattack}
  \begin{tabular}{lcc} 
    \hline
    Method   & No Defense & Defense  \\ 
    \hline
    Clean    & 100.00 & 86.12   \\
    FGSM~\cite{Goodfellow2014ExplainingAH}     & 53.70  & 86.12   \\
    PGD~\cite{Madry2017TowardsDL}      & 0.74   & 73.78   \\
    MI~\cite{Dong2017BoostingAA}   & 0.42   & 62.04   \\
    DI~\cite{Xie2018ImprovingTO}   & 2.24   & 62.28   \\
    MD~\cite{Xie2018ImprovingTO}   & 2.26   & 47.46   \\
    A-ce~\cite{Croce2020ReliableEO}  & 0.56   & 72.64   \\
    A-dlr~\cite{Croce2020ReliableEO} & 0.36   & 72.42   \\
  \hline
  \end{tabular}
  \end{table}

\section{Conclusion}

In this paper, we demonstrate the hazards of residual adversarial perturbations and advocate for adversarial purification methods that eliminates adversarial perturbations wherever possible. Based on this theory, we propose a novel adversarial purification method MIPure that defend against adversarial attacks by maximizing the elimination of same-position perturbations and resisting content-similar perturbations. We construct RIAE to constrain the scale of the same-position perturbations on the purified image by masking the image patch information to destroy the perturbatiosn and reconstruct the patch. Then we also propose RCM to encourage our model to resist the influence of content-similar perturbations by simulating residual perturbations. Finally, we propose a joint constraint on pixel loss and perceptual loss to make the generation of purified images more flexible. Experiments show that our adversarial purification model is very effective in defending against adversarial attacks and also exhibits good robustness in the face of agnostic attacks. In future work, we will continue to optimize the structure of the autoencoder and solve problems such as training instability and adversarial sample augmentation. 

\section*{Acknowledgments}

This work was supported in part by the National Natural Science Foundation of China under Grant 61871226, Grant 61571230, Grant 61802190, and Grant 61906093; in part by the Jiangsu Provincial Social Developing Project under Grant BE2018727; and in part by the Open Research Fund in 2021 of Jiangsu Key Laboratory of Spectral Imaging and Intelligent Sense under Grant JSGP202101 and Grant JSGP202204.









\bibliographystyle{elsarticle-harv}
\bibliography{egbib}

\end{document}